\documentclass[10pt,twocolumn,letterpaper]{article}

\usepackage[pagenumbers]{cvpr} %

\usepackage[utf8]{inputenc} %
\usepackage[T1]{fontenc}    %
\usepackage{url}            %
\usepackage{booktabs}       %
\usepackage{amsfonts}       %
\usepackage{nicefrac}       %
\usepackage{microtype}      %
\usepackage{graphicx}
\usepackage{booktabs}
\usepackage[normalem]{ulem}
\usepackage{multirow}
\usepackage{amsmath}
\usepackage{rotating}
\usepackage{makecell}
\usepackage{stfloats}
\usepackage{wrapfig}
\usepackage{enumitem}
\usepackage{scalerel}
\usepackage{color}
\usepackage{gensymb}
\usepackage{bm}
\usepackage[dvipsnames]{xcolor}
\usepackage{scalerel}

\definecolor{cvprblue}{rgb}{0.21,0.49,0.74}
\usepackage[pagebackref,breaklinks,colorlinks,citecolor=cvprblue]{hyperref}

\DeclareFontFamily{U}{mathb}{}
\DeclareFontShape{U}{mathb}{m}{n}{
  <-5.5> mathb5
  <5.5-6.5> mathb6
  <6.5-7.5> mathb7
  <7.5-8.5> mathb8
  <8.5-9.5> mathb9
  <9.5-11.5> mathb10
  <11.5-> mathb12
}{}
\DeclareSymbolFont{mathb}{U}{mathb}{m}{n}
\DeclareMathSymbol{\drsh}{3}{mathb}{"EB}

\newcommand\blfootnote[1]{%
  \begingroup
  \renewcommand\thefootnote{}\footnote{#1}%
  \addtocounter{footnote}{-1}%
  \endgroup
}

\definecolor{citecolor}{HTML}{0071bc}
\definecolor{gtred}{HTML}{FF3E30}
\definecolor{predblue}{HTML}{0776FF}
\hypersetup{
  breaklinks   = true,
  colorlinks   = true, %
  urlcolor     = RubineRed, %
  linkcolor    = RubineRed, %
  citecolor    = citecolor %
}

\newcommand{\tocite}[1]{\textcolor{red}{[TOCITE]}}

\newcommand{\PAR}[1]{\vskip4pt \noindent{\bf #1~}}

\def\paperID{11844} %
\def\confName{CVPR}
\def\confYear{2024}

\title{Efficient LoFTR: Semi-Dense Local Feature Matching with Sparse-Like Speed}

\author{
    \quad Yifan Wang$^{*}$ 
    \quad Xingyi He$^{*}$
    \quad Sida Peng 
    \quad Dongli Tan
    \quad Xiaowei Zhou$^{\dagger}$
    \vspace{1em}
    \\
    Zhejiang University \quad 
}

\begin{document}
\maketitle
\blfootnote{$^*$Equal contribution. The authors from Zhejiang University are affiliated with the State Key Lab of CAD\&CG. $^\dagger$Corresponding author: Xiaowei Zhou.}
\begin{abstract}
We present a novel method for efficiently producing semi-dense matches across images.
Previous detector-free matcher LoFTR has shown remarkable matching capability in handling large-viewpoint change and texture-poor scenarios but suffers from low efficiency.
We revisit its design choices and derive multiple improvements for both efficiency and accuracy.
One key observation is that performing the transformer over the entire feature map is redundant due to shared local information, therefore we propose an aggregated attention mechanism with adaptive token selection for efficiency.
Furthermore, we find spatial variance exists in LoFTR's fine correlation module, which is adverse to matching accuracy.
A novel two-stage correlation layer is proposed to achieve accurate subpixel correspondences for accuracy improvement.
Our efficiency optimized model is $\sim 2.5\times$ faster than LoFTR which can even surpass state-of-the-art efficient sparse matching pipeline SuperPoint + LightGlue. 
Moreover, extensive experiments show that our method can achieve higher accuracy compared with competitive semi-dense matchers, with considerable efficiency benefits.
This opens up exciting prospects for large-scale or latency-sensitive applications such as image retrieval and 3D reconstruction.
Project page: \url{https://zju3dv.github.io/efficientloftr/}.
\end{abstract}
\section{Introduction}
\label{sec:intro}

Image matching is the cornerstone of many 3D computer vision tasks, which aim to find a set of highly accurate correspondences given an image pair.
The established matches between images have broad usages such as reconstructing the 3D world by structure from motion~(SfM)~\cite{Agarwal2009BuildingRI,schonbergerStructurefromMotionRevisited2016,Lindenberger2021PixelPerfectSW,he2023detectorfree} or SLAM system~\cite{MurArtal2015ORBSLAMAV,MurArtal2016ORBSLAM2AO}, and visual localization~\cite{Sarlin2018FromCT,sarlin21pixloc}, etc.
Previous methods typically follow a two-stage pipeline: they first detect~\cite{rosten2006machine} and describe~\cite{tian2017l2} a set of keypoints on each image, and then establish keypoint correspondences by handcrafted~\cite{LoweDavid2004DistinctiveIF} or learning-based matchers~\cite{sarlin20superglue,lindenberger2023lightglue}.
These detector-based methods are efficient but suffer from robustly detecting repeatable keypoints across challenging pairs, such as extreme viewpoint changes and texture-poor regions.

\begin{figure}[tp]
    \centering
    \includegraphics[width=1.0\linewidth]{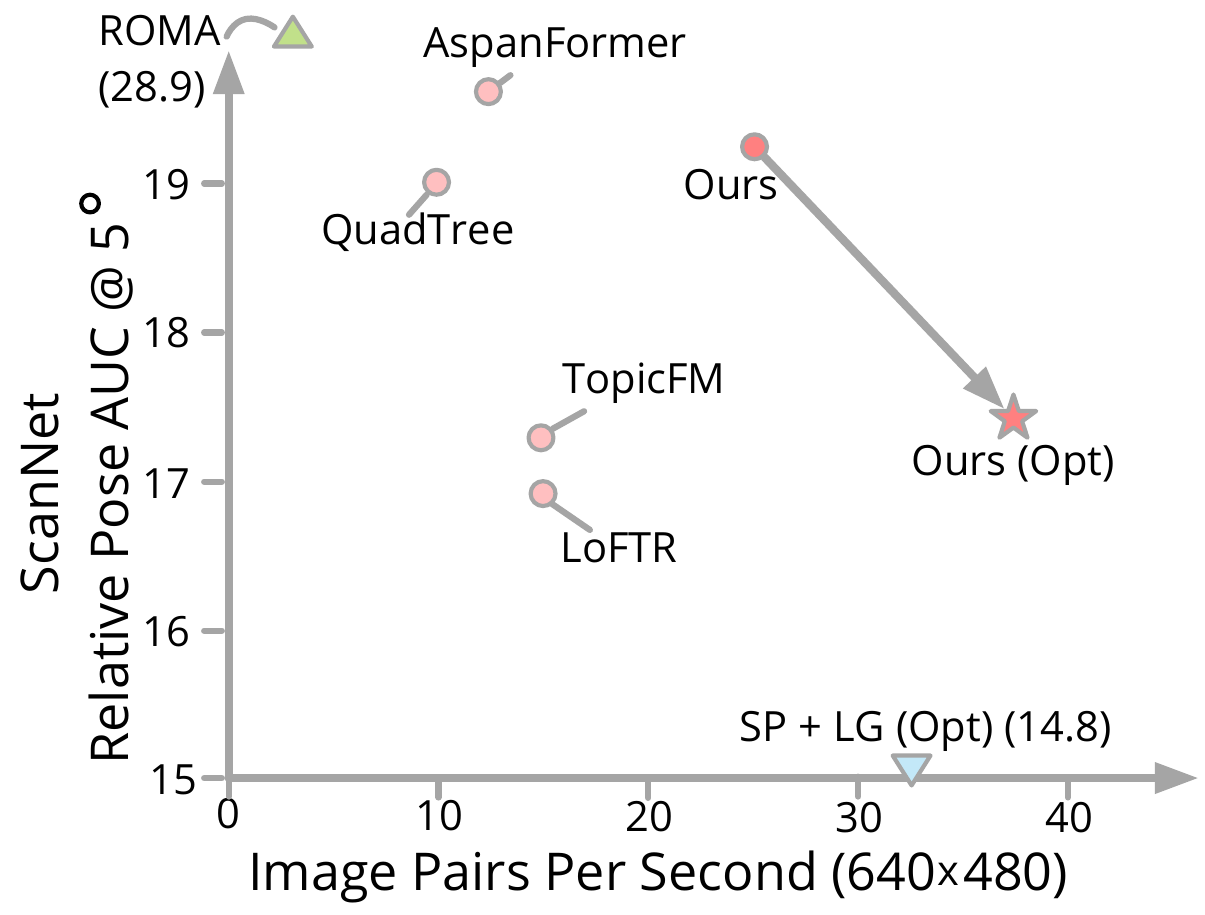}
    \vspace{-0.7 cm}
    \caption{\textbf{Matching Accuracy and Efficiency Comparisons.} 
    Our method achieves competitive accuracy compared with semi-dense matchers~(\protect\scalerel*{\includegraphics{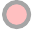}}{B}) at a significantly higher speed.
    Compared with dense matcher ROMA~(\protect\scalerel*{\includegraphics{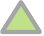}}{B}), our method is $\sim 7.5\times$ faster.
    Moreover, our efficiency optimized model~(\protect\scalerel*{\includegraphics{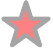}}{B}) can surpass the robust sparse matching pipeline~(\protect\scalerel*{\includegraphics{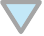}}{B}) SuperPoint~(SP) + LightGlue~(LG) on efficiency with considerably better accuracy.
    }
    \vspace{-0.55 cm}
    \label{fig:teaser}
\end{figure}

Recently, LoFTR~\cite{sun2021loftr} introduces a detector-free matching paradigm with transformer to directly establish semi-dense correspondences between two images without detecting keypoints.
With the help of the transformer mechanism to capture the global image context and the detector-free design, LoFTR shows a strong capability of matching challenging pairs, especially in texture-poor scenarios.
To reduce the computation burden, LoFTR adopts a coarse-to-fine pipeline by first performing dense matching on downsampled coarse features maps, where transformer is applied.
Then, the feature locations of coarse matches on one image are fixed, while their subpixel correspondences are searched on the other image by cropping feature patches based on coarse match, performing the feature correlation, and calculating expectation over the correlation patch.

Despite its impressive matching performance, LoFTR suffers from limited efficiency due to the large token size of performing transformer on the entire coarse feature map, which significantly barricades practical large-scale usages such as image retrieval~\cite{Hausler2021PatchNetVLADMF} and SfM~\cite{schonbergerStructurefromMotionRevisited2016}.
A large bunch of LoFTR's follow-up works~\cite{wang2022matchformer,chen2022aspanformer,tang2022quadtree,giang2022topicfm,Ni2023PATSPA} have attempted to improve its matching accuracy.
However, there are rare methods that focus on matching efficiency of detector-free matching.
QuadTree Attention~\cite{tang2022quadtree} incorporates multi-scale transformation with a gradually narrowed attention span to avoid performing attention on large feature maps.
This strategy can reduce the computation cost, but it also divides a single coarse attention process into multiple steps, leading to increased latency.

In this paper, we revisit the design decisions of the detector-free matcher LoFTR, and propose a new matching algorithm that squeezes out redundant computations for significantly better efficiency while further improving the accuracy.
As shown in Fig.~\ref{fig:teaser}, our approach achieves the best inference speed compared with recent image matching methods while being competitive in terms of accuracy.
Our key innovations lie in introducing a token aggregation mechanism for efficient feature transformation and a two-stage correlation layer for correspondence refinement.
Specifically, we find that densely performing global attention over the entire coarse feature map as in LoFTR is unnecessary, as the attention information is similar and shared in the local region.
Therefore, we devise an aggregated attention mechanism to perform feature transformation on adaptively selected tokens, which is significantly compact and effectively reduces the cost of local feature transformation.

In addition, we observe that there can be spatial variance in the matching refinement phase of LoFTR, which is caused by the expectation over the entire correlation patch when noisy feature correlation exists.
To solve this issue, our approach designs a two-stage correlation layer that first locates pixel-level matches with the accurate mutual-nearest-neighbor matching on fine feature patches, and then further refines matches for subpixel-level by performing the correlation and expectation locally within tiny patches.

Extensive experiments are conducted on multiple tasks, including homography estimation, relative pose recovery, as well as visual localization, to show the efficacy of our method.
Our pipeline pushes detector-free matching to unprecedented efficiency, which is $\sim2.5$ times faster than LoFTR and can even surpass the current state-of-the-art efficient sparse matcher LightGlue~\cite{lindenberger2023lightglue}.
Moreover, our framework can achieve comparable or even better matching accuracy compared with competitive detector-free baselines~\cite{chen2022aspanformer,edstedt2023dkm,edstedt2023roma} with considerably higher efficiency.

In summary, this paper has the following contributions:
\begin{itemize}

\item A new detector-free matching pipeline with multiple improvements based on the comprehensive revisiting of LoFTR, which is significantly more efficient and with better accuracy.
\item A novel aggregated attention network for efficient local feature transformation.
\item A novel two-stage correlation refinement layer for accurate and subpixel-level refined correspondences.
\end{itemize}

\section{Related Work}

\paragraph{Detector-Based Image Matching.}
Classical image matching methods~\cite{LoweDavid2004DistinctiveIF, rosten2006machine, bay2008speeded} adopt handcrafted critics for detecting keypoints, describing and then matching them.
Recent methods draw benefits from deep neural networks for both detection~\cite{rosten2006machine,savinov2017quad,barroso2019key} and description~\cite{tian2017l2,mishchuk2017working,tian2019sosnet,ebel2019beyond}, where the robustness and discriminativeness of local descriptors are significantly improved.
Besides, some methods ~\cite{dusmanu2019d2,DeTone2017SuperPointSI,revaud2019r2d2,luo2020aslfeat,tian2020d2d} managed to learn the detector and descriptor together. 
SuperGlue~\cite{sarlin20superglue} is a pioneering method that first introduces the transformer mechanism into matching, which has shown notable improvements over classical handcrafted matchers.
As a side effect, it also costs more time, especially with many keypoints to match.
To improve the efficiency, some subsequent works, such as ~\cite{chen2021learning,shi2022clustergnn}, endeavor to reduce the size of the attention mechanism, albeit at the cost of sacrificing performance.
LightGlue~\cite{lindenberger2023lightglue} introduces a new scheme for efficient sparse matching that is adaptive to the matching difficulty, where the attention process can be stopped earlier for easy pairs.
It is faster than SuperGlue and can achieve competitive performance.
However, robustly detecting keypoints across images is still challenging, especially for texture-poor regions.
Unlike them, our method focuses on the efficiency of the detector-free method, which eliminates the restriction of keypoint detection and shows superior performance for challenging pairs.

\PAR{Detector-Free Image Matching.} 
Detector-free methods directly match images instead of relying on a set of detected keypoints, producing semi-dense or dense matches.
NC-Net~\cite{rocco2018neighbourhood} represents all features and possible matches as a 4D correlation volume. 
Sparse NC-Net~\cite{rocco2020efficient} utilizes sparse correlation layers to ease resolution limitations.
Subsequently, DRC-Net~\cite{li20dualrc} improves efficiency and further improves performance in a coarse-to-fine manner.

LoFTR~\cite{sun2021loftr} first employs the Transformer in detector-free matching to model the long-range dependencies.
It shows remarkable matching capabilities, however, suffers from low efficiency due to the huge computation of densely transforming entire coarse feature maps.
Many follow-up works further improve the matching accuracy.
Matchformer~\cite{wang2022matchformer} and AspanFormer~\cite{chen2022aspanformer} perform attention on multi-scale features, where local attention regions of \cite{chen2022aspanformer} are found with the help of estimated flow.
QuadTree~\cite{tang2022quadtree} gradually restricts the attention span during hierarchical attention to relevant areas, which can reduce overall computation.
However, these designs contribute marginally or even decrease efficiency, since the hierarchical nature of multi-scale attention will further introduce latencies.
TopicFM~\cite{giang2022topicfm} first assigns features with similar semantic meanings to the same topic, where attention is conducted within each topic for efficiency.
Since it needs to sequentially process each token's features for transformation, the efficiency improvement is limited.
Moreover, performing local attention within topics can potentially restrict the capability of modeling long-range dependencies.
Compared with them, the proposed aggregated attention module in our method significantly improves efficiency while achieving better accuracy.

Dense matching methods~\cite{truong2021learning,edstedt2023dkm,edstedt2023roma} are designed to estimate all possible correspondences between two images, which show strong robustness.
However, they are generally much slower compared with sparse and semi-dense methods.
Unlike them, our method produces semi-dense matches with competitive performance and considerably better efficiency.

\PAR{Transformer} has been broadly used in multiple vision tasks, including feature matching.
The efficiency and memory footprint of handling large token sizes are the main limitations of transformer~\cite{Vaswani2017AttentionIA}, where some methods~\cite{Wang2020LinformerSW,Katharopoulos2020TransformersAR,Kitaev2020ReformerTE} attempt to reduce the complexity to a linear scale to alleviate these problems. 
Some methods~\cite{xFormers2022,dao2022flashattention} propose optimizing transformer models for specific hardware architectures for memory and running-time efficiency.
They are orthogonal to our method and can be naturally adapted into the pipeline for further efficiency improvement.

\begin{figure*}[tp]
    \centering
    \includegraphics[width=1.0\linewidth]{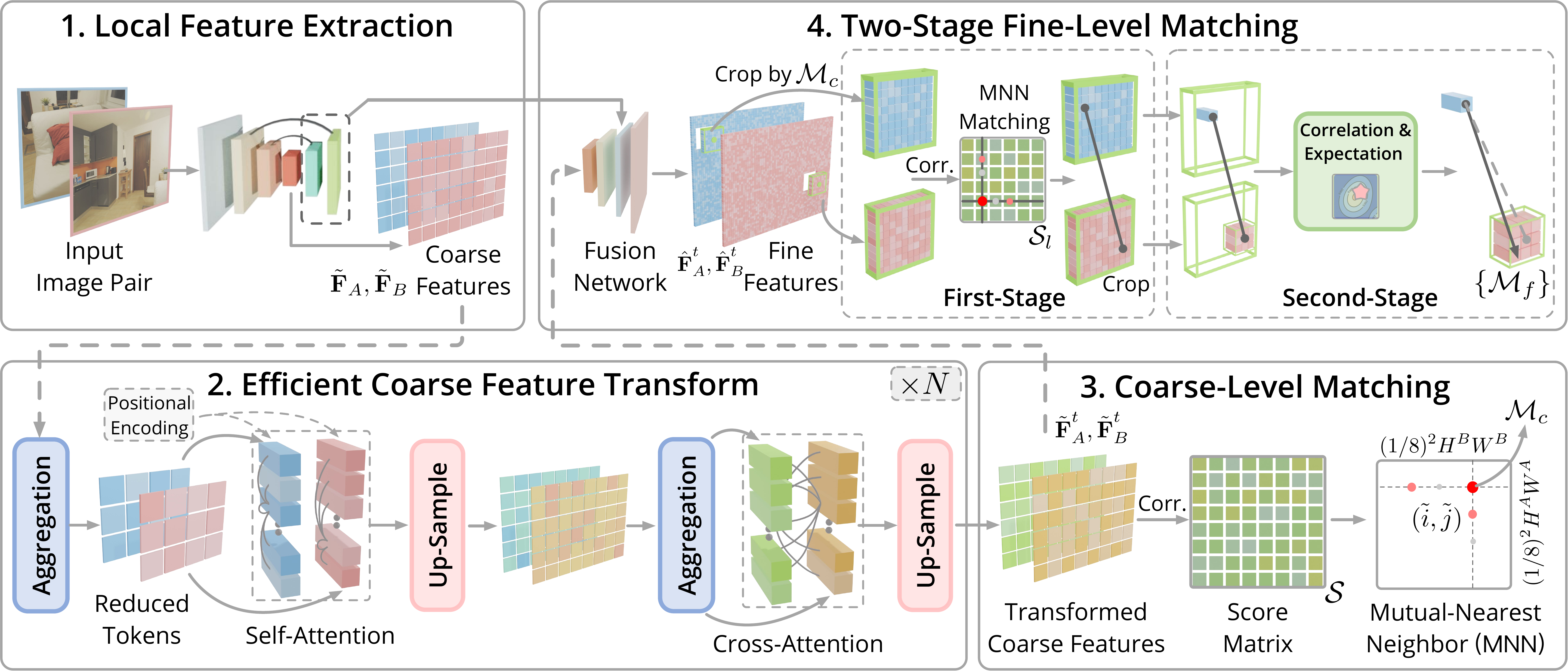}
    \vspace{-0.25 cm}
    \caption{\textbf{Pipeline Overview.} 
    \textbf{(1)} Given an image pair, a CNN network extracts coarse feature maps $\tilde{\textbf{F}}_A$ and $\tilde{\textbf{F}}_B$, as well as fine features.
    \textbf{(2)} Then, we transform coarse features for more discriminative feature maps by interleaving our aggregated self- and cross-attention $N$ times, where adaptively feature aggregation is performed to reduce token size before each attention for efficiency.
    \textbf{(3)} Transformed coarse features are correlated for the score matrix $\mathcal{S}$. Mutual-nearest-neighbor~(MNN) searching is followed to establish coarse matches $\{\mathcal{M}_c\}$.
    \textbf{(4)} To refine coarse matches, discriminative fine features $\hat{\textbf{F}}_A^t$, $\hat{\textbf{F}}_B^t$ in full resolution are obtained by fusing transformed coarse features $\tilde{\textbf{F}}_A^t$, $\tilde{\textbf{F}}_B^t$ with backbone features.
    Feature patches are then cropped centered at each coarse match $\mathcal{M}_c$.
    A two-stage refinement is followed to obtain sub-pixel correspondence $\mathcal{M}_f$.
    }
    \label{fig:allpipeline}
    \vspace{-0.35 cm}
\end{figure*}

\section{Method}

Given a pair of images $\*I_A, \*I_B$, our objective is to establish a set of reliable correspondences between them. 
We achieve this by a coarse-to-fine matching pipeline, which first establishes coarse matches on downsampled feature maps and then refines them for high accuracy.
An overview of our pipeline is shown in Fig.~\ref{fig:allpipeline}.

\subsection{Local Feature Extraction}

Image feature maps are first extracted by a lightweight backbone for later transformation and matching.
Unlike LoFTR and many other detector-free matchers that use a heavy multi-branch ResNet~\cite{He2015DeepRL} network for feature extraction, we alternate to a lightweight single-branch network with reparameterization~\cite{ding2021repvgg} to achieve better inference efficiency while preserving the model performance.

In particular, a multi-branch CNN network with residual connections is applied during training for maximum representational power.
At inference time, we losslessly convert the feature backbone into an efficient single-branch network by adopting the reparameterization technique~\cite{ding2021repvgg}, which is achieved by fusing parallel convolution kernels into a single one.
Then, the intermediate $\nicefrac{1}{8}$ down-sampled coarse features $\tilde{\textbf{F}}_A$, $\tilde{\textbf{F}}_B$ and fine features in $\nicefrac{1}{4}$ and $\nicefrac{1}{2}$ resolutions are extracted efficiently for later coarse-to-fine matching.

\subsection{Efficient Local Feature Transformation}

After the feature extraction, the coarse-level feature maps $\tilde{\textbf{F}}_A$ and $\tilde{\textbf{F}}_B$ are transformed by interleaving self- and cross-attention \footnote{We feed feature of one image as query and feature of the other image as key and value into cross-attention, similar to SG~\cite{sarlin20superglue} and LoFTR~\cite{sun2021loftr}.}
$n$ times to improve discriminativeness.
The transformed features are denoted as $\tilde{\textbf{F}}_A^t$, $\tilde{\textbf{F}}_B^t$.

Previous methods often perform attention on the entire coarse-level feature maps, where linear attention instead of vanilla attention is applied to ensure a manageable computation cost.
However, the efficiency is still limited due to the large token size of coarse features.
Moreover, the usage of linear attention leads to sub-optimal model capability.
Unlike them, we propose efficient aggregated attention for both efficiency and performance.

\paragraph{Preliminaries.}
First, we provide a brief overview of the commonly used vanilla attention and linear attention.
Vanilla attention is a core mechanism in transformer encoder layer, relying on three inputs: query Q, key K, and value V. 
The resultant output is a weighted sum of the value, where the weighted matrix is determined by the query and its corresponding key. 
Formally, the attention function is defined as follows:
\begin{equation}
    \operatorname{VanillaAttention}(Q,K,V)=\operatorname{softmax}({QK^T})V \enspace.
\end{equation}
However, applying the vanilla attention directly to dense local features is impractical due to the significant token size.
To address this issue, previous methods use linear attention to reduce the computational complexity from quadratic to linear:
\begin{equation}
    \operatorname{LinearAttention}(Q,K,V)=\phi(Q)(\phi(K)^T\phi(V)) \enspace.
\end{equation}
where $\phi(\cdot)=elu(\cdot)+1$. 
However, it comes at the cost of reduced representational power, which is also observed by~\cite{cai2022efficientvit}.

\begin{figure}[tp]
    \vspace{-0.5 cm}
    \centering
    \includegraphics[width=\linewidth]{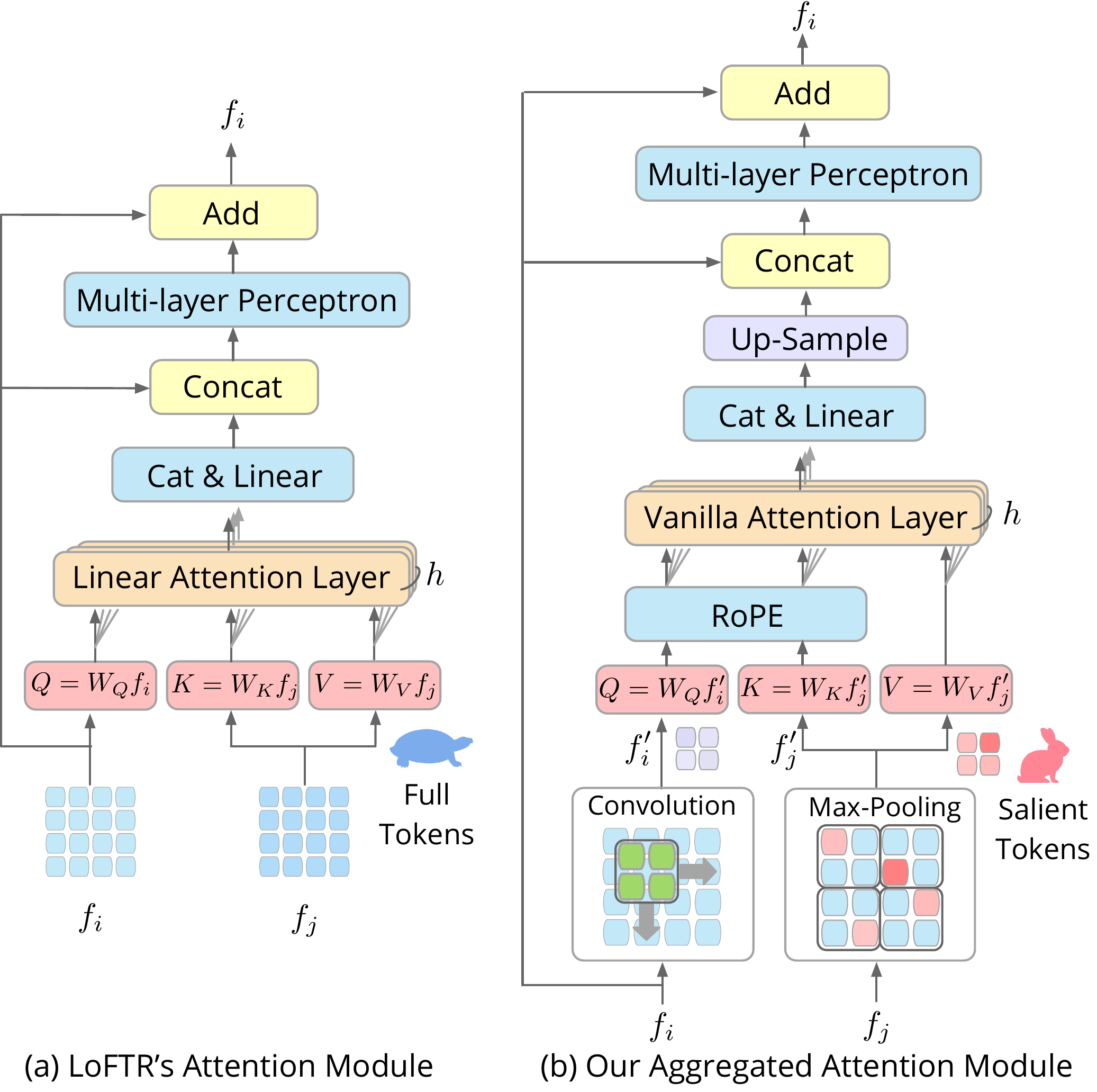}
    \caption{\textbf{Detailed Transformer Module Comparison.}
    Unlike LoFTR which uses all tokens of feature maps to compute attention and resort to linear attention to reduce the computational cost, the proposed attention module first aggregates features for salient tokens, which is significantly more efficient for attention.
    Then the vanilla attention is utilized to transform aggregated features, where relative positional encoding is inserted to capture the spatial information.
    Transformed features are upsampled and fused with the original features to form the final features.
    }
    \vspace{-0.55 cm}
    \label{fig:networkcomparison}
\end{figure}

\paragraph{Aggragated Attention Module.}
After comprehensively investigating the mechanism of the transformer on coarse feature maps, we have two observations that motivate us to devise a new efficient aggregated attention.
First, the attention regions of neighboring query tokens are similar, thus we can aggregate the neighboring tokens of $f_i$ to prevent the redundant computation.
Second, most of the attention weights of each query token are concentrated on a small number of key tokens, hence we can select the salient tokens of $f_j$ before attention to reduce the computation.

Therefore, we propose to first aggregate the $f_i$ token utilizing a depth-wise convolution network, and $f_j$ is aggregated by a max pooling layer to get reduced salient tokens:
\begin{equation}
    f'_i=\operatorname{Conv2D}(f_i),\enspace f'_j=\operatorname{MaxPool}(f_j) \enspace,
\end{equation}
where $\operatorname{Conv2D}$ is implemented by a strided depthwise convolution with a kernel size of $s \times s$, identical to that of the max-pooling layer.
Then positional encoding and vanilla attention are followed to process reduced tokens.
Positional encoding~(PE) can help to model the spatial location contexts, 
where RoPE~\cite{Su2021RoFormerET} is adopted in practice to account for more robust relative positions, inspired by~\cite{lindenberger2023lightglue}.
Note that the PE layer is enabled exclusively for self-attention and skipped during cross-attention.
The transformed feature map is then upsampled and fused with $f_i$ for the final feature map.
Due to the aggregation and selection, the number of tokens in $f'_i$ and $f'_j$ is reduced by $s^2$, which contributes to the efficiency of the attention phase.

\subsection{Coarse-level Matching Module}

We establish coarse-level matches based on the previously transformed coarse feature maps $\tilde{\textbf{F}}_A^t$, $\tilde{\textbf{F}}_B^t$.
Coarse correspondences indicate rough match regions for later subpixel-level matching in the refinement phase.
To achieve this, $\tilde{\textbf{F}}_A^t$ and $\tilde{\textbf{F}}_B^t$ are densely correlated to build a score matrix $\mathcal{S}$.
The softmax operator on both $\mathcal{S}$ dimensions (referred to as dual-softmax) is then applied to obtain the probability of mutual nearest matching, which is commonly used in ~\cite{rocco2018neighbourhood, tyszkiewicz2020disk, sun2021loftr}.
The coarse correspondences $\{\mathcal{M}_c\}$ are established by selecting matches above the score threshold $\tau$ while satisfying the mutual-nearest-neighbor (MNN) constraint.

\paragraph{Efficient Inference Strategy.}
We observe that the dual-softmax operator in the coarse matching can significantly restrict the efficiency in inference due to the large token size, especially for high-resolution images.
Moreover, we find that the dual-softmax operator is crucial for training, dropping it at inference time while directly using the score matrix $\mathcal{S}$ for MNN matching can also work well with better efficiency.

The reason for using the dual-softmax operator in training is that it can help to train discriminative features.
Intuitively, with the softmax operation, the matching score between two pixels can also conditioned on other pixels.
This mechanism forces the network to improve feature similarity of true correspondences while suppressing similarity with irrelevant points.
With trained discriminative features, the softmax operation can be potentially eliminated during inference.

We denote the model skipping dual-softmax layer in inference as \emph{efficiency optimized model}.
Results in Tab.~\ref{tab:exp relativepose} demonstrate the effectiveness of this design.

\subsection{Subpixel-Level Refinement Module}
As overviewed in Fig.~\ref{fig:allpipeline}~(4), with established coarse matches ${\{\mathcal{M}_c\}}$, we refine them for sub-pixel accuracy with our refinement module.
It is composed of an efficient feature patch extractor for discriminative fine features, followed by a two-stage feature correlation layer for final matches ${\{\mathcal{M}_f\}}$.

\paragraph{Efficient Fine Feature Extraction.}
We first extract discriminative fine feature patches centered at each coarse match $\mathcal{M}_c$ by an efficient fusion network for later match refinement.
For efficiency, our key idea here is to re-leverage the previously transformed coarse features $\tilde{\textbf{F}}_A^t$, $\tilde{\textbf{F}}_B^t$ to obtain cross-view attended discriminative fine features, instead of introducing additional feature transform networks as in LoFTR~\cite{sun2021loftr}.

To be specific, $\tilde{\textbf{F}}_A^t$ and $\tilde{\textbf{F}}_B^t$ are adaptively fused with $\nicefrac{1}{4}$ and $\nicefrac{1}{2}$ resolution backbone features by convolution and upsampling to obtain fine feature maps $\hat{\textbf{F}}_A^t$, $\hat{\textbf{F}}_B^t$ in the original image resolution.
Then local feature patches are cropped on fine feature maps centered at each coarse match.
Since only shallow feed-forward networks are included, our fine feature fusion network is remarkably efficient.

\paragraph{Two-Stage Correlation for Refinement.}
Based on the extracted fine local feature patches of coarse matches, we search for high-accurate sub-pixel matches.
To refine a coarse match, a commonly used strategy~\cite{sun2021loftr,chen2022aspanformer,giang2022topicfm} is to select the center-patch feature of $\*I_A$ as a fixed reference point, and perform feature correlation and expectation on the entire corresponding feature patch for its fine match.
However, this refinement-by-expectations will introduce location variance to the final match, because irrelevant regions also have weights and can affect results.
Therefore, we propose a novel two-stage correlation module to solve this problem.

Our idea is to utilize a mutual-nearest-neighbor~(MNN) matching to get intermediate pixel-level refined matches in the first stage, and then refine them for subpixel accuracy by correlation and expectation.
Motivations are that MNN matching don't have spatial variance since matches are selected by directly indexing pixels with maximum scores, but cannot achieve sub-pixel accuracy.
Conversely, refinement-by-expectation can achieve sub-pixel accuracy but variance exists.
The proposed two-stage refinement can draw benefits by combining the best of both worlds.

Specifically, to refine a coarse-level correspondence $\mathcal{M}_c$, the first-stage refinement phase densely correlates their fine feature patches to obtain the local patch score matrix $\mathcal{S}_l$.
MNN searching is then applied on $\mathcal{S}_l$ to get intermediate pixel-level fine matches.
To limit the overall match number, we select the top-$1$ fine match for one coarse match by sorting the correlation scores.

Then, we further refine these pixel-level matches for subpixel accuracy by our second-stage refinement.
Since the matching accuracy has already significantly improved in first-stage refinement, now we can use a tiny local window for correlation and expectation with a maximum suppression of location variance.
In practice, we correlate the feature of each point in $\textbf{I}_A$ with a $3\times3$ feature patch centered at its fine match in $\textbf{I}_B$.
The softmax operator is then applied to get a match distribution matrix and the final refined match is obtained by calculating expectations.

\subsection{Supervision}
The entire pipeline is trained end-to-end by supervising the coarse and refinement matching modules separately.

\paragraph{Coarse-Level Matching Supervision.}
The coarse ground truth matches $\{\mathcal{M}_c\}_{gt}$ with a total number of $N$ are built by warping grid-level points from $\textbf{I}_A$ to $\textbf{I}_B$ via depth maps and image poses following previous methods~\cite{sarlin20superglue,sun2021loftr}.
The produced correlation score matrix $\mathcal{S}$ in coarse matching is supervised by minimizing the log-likelihood loss over locations of $\{\mathcal{M}_c\}_{gt}$: %
\begin{equation}
\resizebox{0.75\columnwidth}{!}{
    $
    \mathcal{L}_{c} = - \frac{1}{N}
    \sum_{(\tilde{i}, \tilde{j}) \in \{\mathcal{M}_{c}\}_{gt}} \log \mathcal{S}\left(\tilde{i}, \tilde{j}\right) \enspace.
    $
}
    \label{eq:loss_coarse}
\end{equation}

\paragraph{Fine-Level Matching Supervision.}
We train the proposed two-stage fine-level matching module by separately supervising the two phases.
The first stage fine loss $\mathcal{L}_{f1}$ is to minimize the log-likelihood loss of each fine local score matrix $\mathcal{S}_l$ based on the pixel-level ground truth fine matches, similar to coarse loss.
The second stage is trained by $\mathcal{L}_{f2}$ that calculates the $\ell_2$ loss between the final subpixel matches $\{\mathcal{M}_f\}$ and ground truth fine matches $\{\mathcal{M}_f\}_{gt}$.

The total loss is the weighted sum of all supervisions: $\mathcal{L} = \mathcal{L}_c + \alpha \mathcal{L}_{f1} + \beta \mathcal{L}_{f2}$.
\section{Experiments}
In this section, we evaluate the performance of our method on several downstream tasks, including homography estimation, pairwise pose estimation and visual localization. Furthermore, we evaluate the effectiveness of our design by conducting detailed ablation studies.
\begin{table*}[t]
    \vspace{-0.4 cm}
    \centering
    \resizebox{0.9\textwidth}{!}{
    \setlength\tabcolsep{10pt} %
    \begin{tabular}{ccccccccc} 
    \toprule
    \multirow{2}{*}{Category} & \multirow{2}{*}{Method}         & \multicolumn{3}{c}{MegaDepth Dataset}             & \multicolumn{3}{c}{ScanNet Dataset} & \multirow{2}{*}{Time~(ms)} \\ 
    \cmidrule(lr){3-5}
    \cmidrule(lr){6-8}
        &              & AUC@5$\degree$       & AUC@10$\degree$       & AUC@20$\degree$  & AUC@5$\degree$       & AUC@10$\degree$       & AUC@20$\degree$  & \\ 
    \midrule
    \multirow{3}{*}{Sparse} &   SP + NN & 31.7&46.8&60.1 & 7.5 & 18.6 & 32.1 & \textbf{10.8}\\
    & SP + SG & 49.7&\textbf{67.1}&\textbf{80.6} &  
    \textbf{16.2}&\textbf{32.8}&\textbf{49.7} & 48.3 \\
    & SP + LG & \textbf{49.9}&67.0&80.1 & 14.8&30.8&47.5&31.9/30.7 \\
    \hline
    \multirow{7}{*}{Semi-Dense} & DRC-Net & 27.0&42.9&58.3 &7.7&17.9&30.5 & 328.0 \\
    & LoFTR & 52.8&69.2&81.2 &16.9&33.6&50.6& 66.2\\
    & QuadTree & 54.6&70.5&82.2 &19.0&37.3&53.5 & 100.7\\
    & MatchFormer & 53.3&69.7&81.8 & 15.8&32.0&48.0 &128.9 \\
    & TopicFM & 54.1&70.1&81.6 & 17.3&35.5&50.9 & 66.4 \\
    & AspanFormer & 55.3&71.5&83.1 & \textbf{19.6}&\textbf{37.7}&\textbf{54.4} &81.6  \\
    & Ours & \textbf{56.4}&\textbf{72.2}&\textbf{83.5}& 19.2&37.0&53.6 & 40.1/34.4\\
    & Ours~(Optimized) & 55.4 & 71.4 & 82.9& 17.4& 34.4& 51.2& \textbf{35.6/27.0}\\
    \hline
    \multirow{2}{*}{Dense} & DKM & 60.4 & 74.9 & 85.1 & 26.64 & 47.07 & 64.17 & 210.8\\ %
    & ROMA & 62.6&76.7&86.3 &28.9&50.4&68.3& 302.7\\ %
    \bottomrule
    \end{tabular}
    }
    \vspace{0.15cm}
    \caption{\textbf{Results of Relative Pose Estimation on MegaDepth Dataset and ScanNet Dataset.}
    We use the models trained on the MegaDepth dataset to evaluate all methods on both datasets, which can show the intra- and inter-dataset generalization abilities.
    The AUC of pose error at different thresholds, along with the processing time for matching image pair at a resolution of $640 \times 480$, is presented.
    For SP + LG, Ours, and Ours~(Optimized), the running times of the model using FP32/Mixed-Precision numerical precisions are shown.
    }
    \label{tab:exp relativepose}
    \vspace{-0.35 cm}
    \end{table*}

\subsection{Implementation Details}
We adopt RepVGG~\cite{ding2021repvgg} as our feature backbone, and self- and cross-attention are interleaved for $N=4$ times to transform coarse features.
For each attention, we aggregate features by a depth-wise convolution layer and a max-pooling layer, both with a kernel size of $4\times4$.
Our model is trained on the MegaDepth dataset~\cite{li2018megadepth}, which is a large-scale outdoor dataset.
The test scenes are separated from training data following~\cite{sun2021loftr}. %
The loss function's weights $\alpha$ and $\beta$ are set to $1.0$ and $0.25$, respectively.
We use the AdamW optimizer with an initial learning rate of $4\times10^{-3}$.
The network training takes about 15 hours with a batch size of 16 on $8$ NVIDIA V100 GPUs.
And the coarse and fine stages are trained together from scratch.
The trained model on MegaDepth is used to evaluate all datasets and tasks in our experiments to demonstrate the generalization ability.

\subsection{Relative Pose Estimation}
\PAR{Datasets.} 
We use the outdoor MegaDepth~\cite{li2018megadepth} dataset and indoor ScanNet~\cite{dai2017scannet} dataset for the evaluation of relative pose estimation to demonstrate the efficacy of our method.

MegaDepth dataset is a large-scale dataset containing sparse 3D reconstructions from 196 scenes.
The key challenges on this dataset are large viewpoints and illumination changes, as well as repetitive patterns.
We follow the test split of the previous method~\cite{sun2021loftr} that uses 1500 sampled pairs from scenes ``Sacre Coeur'' and ``St. Peter’s Square'' for evaluation.
Images are resized so that the longest edge equals 1200 for all semi-dense and dense methods.
Following~\cite{lindenberger2023lightglue}, sparse methods are provided resized images with longest edge equals 1600.

ScanNet dataset contains 1613 sequences with ground-truth depth maps and camera poses.
They depict indoor scenes with viewpoint changes and texture-less regions.
We use the sampled test pairs from~\cite{sarlin20superglue} for the evaluation, where images are resized to $640\times480$ for all methods.

\PAR{Baselines.} 
We compare the proposed method with three categories of methods: 1) sparse keypoint detection and matching methods, including SuperPoint~\cite{DeTone2017SuperPointSI} with Nearest-Neighbor~(NN), SuperGlue~(SG)~\cite{sarlin20superglue}, LightGlue~(LG)~\cite{lindenberger2023lightglue} matchers, 2) semi-dense matchers, including DRC-Net~\cite{li20dualrc}, LoFTR~\cite{sun2021loftr}, QuadTree Attention~\cite{tang2022quadtree}, MatchFormer~\cite{wang2022matchformer}, AspanFormer~\cite{chen2022aspanformer}, TopicFM~\cite{giang2022topicfm}, and 3) state-of-the-art dense matcher ROMA~\cite{edstedt2023roma} that predict matches for each pixel.

\PAR{Evaluation protocol.}
Following previous methods, the recovered relative poses by matches are evaluated for reflecting matching accuracy. 
The pose error is defined as the maximum of angular errors in rotation and translation. %
We report the AUC of the pose error at thresholds~(5\degree, 10\degree, and 20\degree).
Moreover, the running time of matching each image pair in the ScanNet dataset is reported for comprehensively understanding the matching accuracy and efficiency balance.
We use a single NVIDIA 3090 to evaluate the running time of all methods.

\begin{figure*}[btp]
	\centering
	\includegraphics[width=0.85\textwidth]{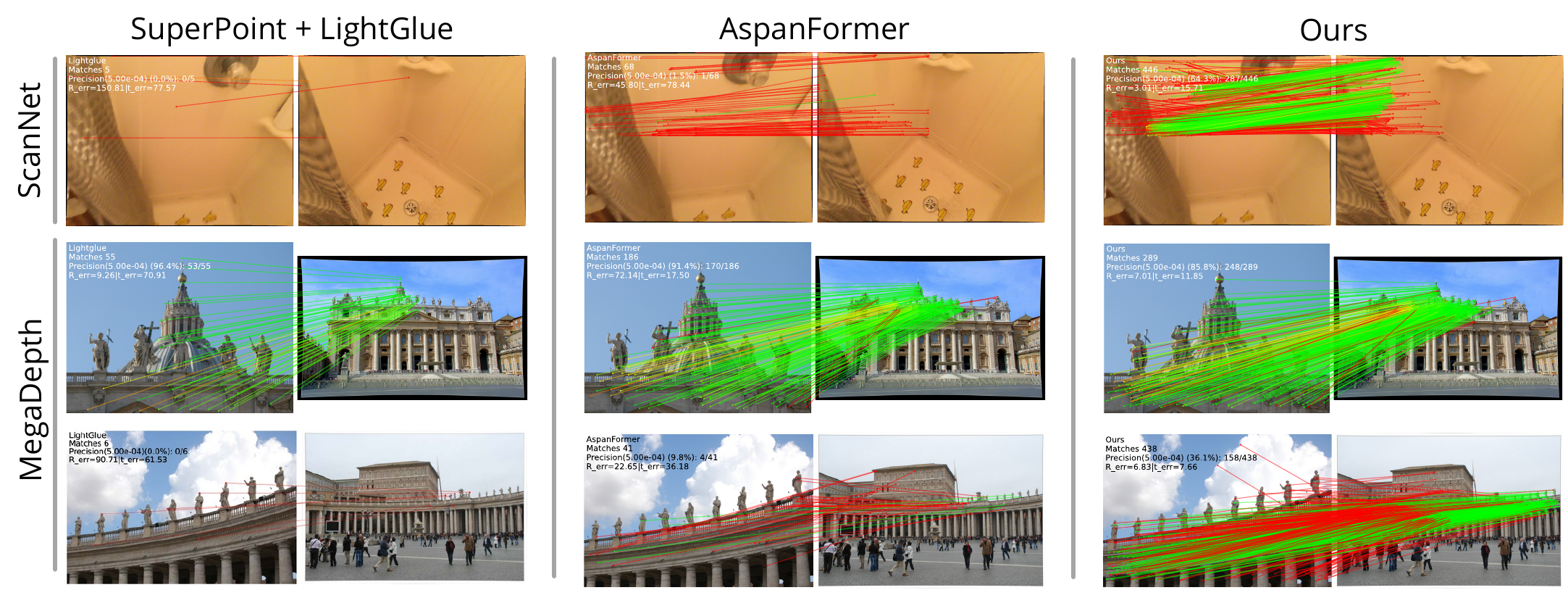}
	\vspace{-0.2 cm}
	\caption{{\textbf{Qualitative Results.}}
	Our method is compared with the sparse matching pipeline SuperPoint~\cite{DeTone2017SuperPointSI}+LightGlue~\cite{lindenberger2023lightglue}, semi-dense matcher AspanFormer~\cite{chen2022aspanformer}. 
	Image pairs with texture-poor regions and large-viewpoint changes can be robustly matched by our method.
	The red color indicates epipolar error beyond $5 \times 10^{-4}$ (in the normalized image coordinates).
	}
	\label{fig:qualitative}
	  \vspace{-0.4 cm}
\end{figure*}

\PAR{Results.}
As shown in Tab.~\ref{tab:exp relativepose}, the proposed method achieves competitive performances compared with sparse and semi-dense methods on both datasets.
Qualitative comparisons are shown in Fig.~\ref{fig:qualitative}.
Specifically, our method outperforms the best semi-dense baseline AspanFormer on all metrics of the MegaDepth dataset and has lower but comparable performance on the ScanNet dataset, with $\sim 2$ times faster.
Our optimized model that eliminates the dual-softmax operator in coarse-level matching further brings efficiency improvements, with slight performance decreases.
Using this strategy, our method can outperform the efficient and robust sparse method SP + LG in efficiency with significantly higher accuracy.
Dense matcher ROMA shows remarkable matching capability but is slow for applications in practice.
Moreover, since ROMA utilizes the pre-trained DINOv2~\cite{oquab2023dinov2} backbone, its strong generalizability on ScanNet may be attributed to the similar indoor training data in DINOv2, where other methods are trained on outdoor MegaDepth only.
Compared with it, our method is $\sim 7.5$ times faster, which has a good balance between accuracy and efficiency.

\begin{table}[t]
    \centering
    \resizebox{1.0\columnwidth}{!}{
    \setlength\tabcolsep{15pt} %
    \begin{tabular}{ccccc} 
    \toprule
    \multirow{2}{*}{Category} & \multirow{2}{*}{Method}         & \multicolumn{3}{c}{Homography est. AUC} \\ 
    \cmidrule(lr){3-5}
        &              & @3px       & @5px       & @10px \\ 
    \midrule
    \multirow{4}{*}{Sparse} &   D2Net + NN & 23.2 & 35.9 & 53.6  \\
    & R2D2 + NN & 50.6 & 63.9 &76.8\\
    & DISK + NN & 52.3 & 64.9 & 78.9\\
    & SP + SG & \textbf{53.9} & \textbf{68.3} & \textbf{81.7} \\
    \hline
    \multirow{5}{*}{Semi-Dense} & Sparse-NCNet & 48.9 & 54.2 & 67.1 \\
    & DRC-Net & 50.6 & 56.2 & 68.3 \\
    & LoFTR & 65.9 & 75.6 & 84.6\\
    & Ours &\textbf{66.5}&\textbf{76.4}&\textbf{85.5} \\
    \bottomrule
    \end{tabular}
    }
    \vspace{-0.15cm}
    \caption{\textbf{Results of Homography Estimation on HPatches Dataset.}
    Our method is compared with sparse and semi-dense methods. 
    The AUC of reprojection error of corner points at different thresholds is reported.
    }
    \label{tab:exp hpatches}
    \vspace{-0.35 cm}
    \end{table}
\subsection{Homography Estimation}
\PAR{Dataset.} We evaluate our method on HPatches dataset~\cite{balntas2017hpatches}.
HPatches dataset depicts planar scenes divided into sequences. 
Images are taken under different viewpoints or illumination changes.

\PAR{Baselines.}
We compare our method with sparse methods including D2Net~\cite{dusmanu2019d2}, R2D2~\cite{revaud2019r2d2}, DISK~\cite{tyszkiewicz2020disk} detectors with NN matcher, and SuperPoint~\cite{DeTone2017SuperPointSI} + SuperGlue~\cite{sarlin20superglue}.
As for semi-dense methods, we compare with Sparse-NCNet~\cite{rocco2020efficient}, DRC-Net~\cite{li20dualrc}, and LoFTR~\cite{sun2021loftr}.
For SuperGlue and all semi-dense methods, we use their models trained on MegaDepth dataset for evaluation.

\PAR{Evaluation Protocol.} Following SuperGlue~\cite{sarlin20superglue} and LoFTR~\cite{sun2021loftr}, we resize all images for matching so that their smallest edge equals 480 pixels. 
We collect the mean reprojection error of corner points, and report the area under the cumulative curve (AUC) under 3 different thresholds, including $3$~px, $5$~px, and $10$~px.
For all baselines, we employ the same RANSAC method as a robust homography estimator for a fair comparison.
Following LoFTR, we select only the top 1000 predicted matches of semi-dense methods for the sake of fairness.

\PAR{Results.}
As shown in Tab.~\ref{tab:exp hpatches}, even though the number of matches is restricted, our method can also work remarkably well and outperform sparse methods significantly.
Compared with semi-dense, our method can surpass them with significantly higher efficiency.
We attribute this to the effectiveness of two-stage refinement for accuracy improvement and proposed aggregation module for efficiency.

\begin{table}[t]
    \centering
    \resizebox{1.0\columnwidth}{!}{
    \setlength\tabcolsep{25pt} %
    \begin{tabular}{ccc} 
    \toprule
    \multirow{2}{*}{Method}         & DUC1 & DUC2\\ 
    \cmidrule(lr){2-3}
        &              \multicolumn{2}{c}{(0.25m,2$\degree$)/(0.5m,5$\degree$)/(1.0m,10$\degree$)} \\ 
    \midrule
    SP+SG & 49.0 / 68.7 / 80.8 & 53.4 / 77.1 / 82.4 \\
    LoFTR & 47.5 / 72.2 / 84.8 & 54.2 / 74.8 / 85.5 \\
    TopicFM & 52.0 / \textbf{74.7} / \textbf{87.4} & 53.4 / 74.8 / 83.2 \\
    PATS & \textbf{55.6} / 71.2 / 81.0 & \textbf{58.8} / \textbf{80.9} / 85.5 \\ 
    AspanFormer & 51.5 / 73.7 / 86.0 & 55.0 / 74.0 / 81.7\\
    Ours & 52.0 / \textbf{74.7} / 86.9 & 58.0 / \textbf{80.9} / \textbf{89.3} \\
    \bottomrule
    \end{tabular}
    }
    \vspace{0.05cm}
    \caption{\textbf{Results of Visual Localization on InLoc Dataset.}
    }
    \label{tab:exp inloc}
    \vspace{-0.35 cm}
    \end{table}
\begin{table}[t]
    \centering
    \resizebox{1.0\columnwidth}{!}{
    \setlength\tabcolsep{25pt} %
    \begin{tabular}{ccc} 
    \toprule
    \multirow{2}{*}{Method}         & Day & Night\\ 
    \cmidrule(lr){2-3}
        &              \multicolumn{2}{c}{(0.25m,2$\degree$)/(0.5m,5$\degree$)/(1.0m,10$\degree$)} \\ 
    \midrule
    SP+SG & 89.8 / 96.1 / \textbf{99.4} & 77.0 / 90.6 / \textbf{100.0} \\
    LoFTR & 88.7 / 95.6 / 99.0 & \textbf{78.5} / 90.6 / 99.0 \\
    TopicFM & \textbf{90.2} / 95.9 / 98.9 & 77.5 / 91.1 / 99.5 \\
    PATS & 89.6 / 95.8 / 99.3 & 73.8 / \textbf{92.1} / 99.5 \\
    AspanFormer & 89.4 / 95.6 / 99.0 & 77.5 / 91.6 / 99.5\\
    Ours & 89.6 / \textbf{96.2} / 99.0 & 77.0 / 91.1 / 99.5\\
    \bottomrule
    \end{tabular}
    }
    \vspace{0.05cm}
    \caption{\textbf{Results of Visual Localization on Aachen v1.1 Dataset.}
    }
    \label{tab:exp aachen}
    \vspace{-0.35 cm}
    \end{table}

\subsection{Visual Localization}
\PAR{Datasets and Evaluation Protocols.}
Visual localization is an important downstream task of image matching, which aims to estimate the 6-DoF poses of query images based on the 3D scene model.
We conduct experiments on two commonly used benchmarks, including InLoc~\cite{taira2018inloc} dataset and Aachen v1.1~\cite{sattler2018benchmarking} dataset, for evaluation to demonstrate the superiority of our method.
InLoc dataset is captured on indoor scenes with plenty of repetitive structures and texture-less regions, where each database image has a corresponding depth map.
Aachen v1.1 is a challenging large-scale outdoor dataset for localization with large-viewpoint and day-and-night illumination changes, which particularly relies on the robustness of matching methods.
We adopt its full localization track for benchmarking.

Following~\cite{sun2021loftr,chen2022aspanformer}, the open-sourced localization framework HLoc~\cite{Sarlin2018FromCT} is utilized.
For both datasets, the percentage of pose errors satisfying both angular and distance thresholds is reported following the benchmarks, where different thresholds are used.
For the InLoc dataset, the metrics of two test scenes including DUC1 and DUC2 are separately reported.
As for the Aachen v1.1 dataset, the metrics corresponding to the daytime and nighttime divisions are reported.

\PAR{Baselines.} 
We compare the proposed method with both detector-based method SuperPoint~\cite{DeTone2017SuperPointSI}+SuperGlue~\cite{sarlin20superglue} and detector-free methods including LoFTR~\cite{sun2021loftr}, TopicFM~\cite{giang2022topicfm}, PATS~\cite{Ni2023PATSPA} and Aspanformer~\cite{chen2022aspanformer}.

\PAR{Results.}
We adhere to the pipeline and evaluation settings of the online visual localization benchmark\footnote{https://www.visuallocalization.net/benchmark} to ensure fairness.
As presented in Tab.~\ref{tab:exp inloc}, the proposed method achieves competitive results, taking both detector-based and detector-free methods into account.
Being a method primarily geared towards efficiency, our approach can deliver results comparable to those of many accuracy-oriented methods.
As depicted in Tab.~\ref{tab:exp aachen}, our method also demonstrates performance on par with the best-performing approaches.

\begin{table}[t]
    \centering
    \resizebox{1.0\columnwidth}{!}{
    \setlength\tabcolsep{2pt} %
    \begin{tabular}{lcccc} 
    \toprule
    \multirow{2}{*}{Method}         & \multicolumn{3}{c}{Pose Estimation AUC} & \multirow{2}{*}{Time~(ms)}\\ 
    \cmidrule(lr){2-4}
        & @5\degree       & @10\degree       & @20\degree \\ 
    \midrule
    Ours Full & \textbf{56.4} & \textbf{72.2} & \textbf{83.5}  & 139.2\\
    1) Ours Optimal~(w/o dual-softmax) & 55.4 & 71.4 & 82.9 & \textbf{102.0} \\ 
    2) Replace Agg. Attention to LoFTR's Trans. & 54.7&70.5&82.2 & 171.4\\
    3) Replace two-stage refine. to LoFTR's refine. & 54.7 & 70.9 & 82.7 & 135.3 \\
    4) No second-stage refinement & 55.8 & 71.8 & 83.3 & 138.1\\
    5) Replace RepVGG with ResNet & 55.4&71.4&82.9 & 156.2\\
    \bottomrule
    \end{tabular}
    }
    \vspace{-0.15cm}
    \caption{\textbf{Ablation Studies.}
    The components of our method are ablated on the MegaDepth dataset for a comprehensive understanding of our method, where averaged running times for an image pair with high-resolution $1200\times1200$ are reported.
    }
    \label{tab:exp ablation}
    \vspace{-0.35 cm}
    \end{table}
\subsection{Ablation Studies}
In this part, we conduct detailed ablation studies to analyze the effectiveness of our proposed modules with results shown in Tab.~\ref{tab:exp ablation}.
1) Without dual-softmax, our optimal model can bring huge efficiency improvement in high-resolution images.
2) In coarse feature transformation, replacing the proposed aggregated attention module with LoFTR's transformer can bring significant efficiency dropping, as well as accuracy decrease. Note that the replaced transformer is also equipped with RoPE same as ours for fair comparison.
This demonstrates the efficacy of the proposed module that performing vanilla attention on aggregated features can achieve higher efficiency with even better matching accuracy.
3) Compared with using LoFTR's refinement that performs expectation on the entire correlation patch, the proposed two-stage refinement layer can bring accuracy improvement with neglectable latency.
We attribute this to the two-stage refinement's property that can maximize the suppression of location variance in correlation refinement.
4) Dropping the second refinement stage will lead to degraded pose accuracy with minor efficiency changes, especially on the strict AUC@$5\degree$ metric.
5) Changing the backbone from reparameterized VGG~\cite{ding2021repvgg} back to multi-branch ResNet~\cite{He2015DeepRL} leads to decreased efficiency with similar accuracy, which demonstrates the effectiveness of our design choice for efficiency.

\section{Conclusions}
This paper introduces a new semi-dense local feature matcher based on the success of LoFTR.
We revisit its designs and propose several improvements for both efficiency and matching accuracy.
A key observation is that performing the Transformer on the entire coarse feature map is redundant due to the similar local information, where an aggregated attention module is proposed to perform transformer on reduced tokens with significantly better efficiency and competitive performance.
Moreover, a two-stage correlation layer is devised to solve the location variance problem in LoFTR's refinement design, which further brings accuracy improvements.
As a result, our method can achieve $\sim 2.5$ times faster compared with LoFTR with better matching accuracy.
Moreover, as a semi-dense matching method, the proposed method can achieve comparable efficiency with the recent robust sparse feature matcher LightGlue~\cite{lindenberger2023lightglue}.
We believe this opens up the applications of our method in large-scale or latency-sensitive downstream tasks, such as image retrieval and 3D reconstruction.
Please refer to the supplementary material for discussions about limitations and future works.

{
   \clearpage
   \small
   \bibliographystyle{ieeenat_fullname}
   \bibliography{egbib}
}
\clearpage
\appendix

\section*{Supplementary Material}
\section{Insight and Discussion about Aggregated Attention Module}
Some previous works explored using pooling in ViT but are with different design choices from our method due to different tasks.
PoolFormer~\cite{Yu2021MetaFormerIA} replaces the multi-head attention with pooling, which cannot be used for cross-attention in matching that two images are not pixel-aligned.
MVit~\cite{Fan2021MultiscaleVT} uses pooling to reduce tokens like ours, but they cannot get high-res features that are required for matching. 

Differently, we propose to first conduct attention on aggregated features and then \emph{upsample} before feed-forward network~(FFN) for later fusion with input feature, as shown in Fig.~3. 
This aggregate-and-upsample block can minimize information loss in aggregation and efficiently get high-res informative features, where conducting upsampling before fusion is crucial to fuse smoothly interpolated messages with a detailed feature map.
Ablation is in Tab.~\ref{tab: abl_scannet}~(8).

Moreover, We perform Conv on Q value instead of pooling because salient tokens should \emph{not} represent neighbors to query attention.
The transformer is crucial for enhancing non-salient features for matching. 
Pooling on Q causes the attention of texture-less areas dominated by neighboring salient tokens, reducing the performance as ablated in Tab.~\ref{tab: abl_scannet}~(6,7).

\section{Implementation Details}

\subsection{Local Feature Extraction}
RepVGG~\cite{ding2021repvgg} blocks are used to build a four-stage feature backbone.
We use a width of 64 and a stride of 1 for the first stage and widths of [64, 128, 256] and strides of 2 for the subsequent three stages.
Each stage is composed of [1, 2, 4, 14] RepVGG blocks and ReLU activations, respectively.
The output of the last stage in \nicefrac{1}{8} image resolution is used for efficient local feature transformer modules to get attended coarse feature maps. 
The second and third stages' feature maps are in \nicefrac{1}{2} and \nicefrac{1}{4} image resolutions, respectively, which are used for fusing with transformed coarse features for fine features.

\subsection{Position Encoding}
We use the 2D extension of Rotary position encoding~\cite{Su2021RoFormerET} to encode the relative position between coarse features
in self-attention modules.
Given the projected features $q$ and $k$, the attention score between two features $q_i$ and $k_j$ is computed as:
\begin{equation}
        a_{ij} = q_i^{T}R(x_j - x_i, y_j - y_i)k_j \enspace,
    \label{eq:score_ij}
\end{equation}
where $x_i, y_i, x_j, y_j$ are the coordinates of $q_i$ and $k_j$, $R$ is a block diagonal matrix:
\begin{equation}
    \resizebox{0.9\columnwidth}{!}{
    $
    R(\Delta{x}, \Delta{y}) = \begin{pmatrix}
        R_1(\Delta{x}, \Delta{y}) &  &  & \\
        & R_2(\Delta{x}, \Delta{y}) &  & \\
        &  &  \ddots & \\
        &  &  & R_{d/4}(\Delta{x}, \Delta{y}) 
    \end{pmatrix}\enspace,
    $
    }
\end{equation}
\begin{equation}
    \resizebox{0.9\columnwidth}{!}{
    $
    R_k(\Delta{x}, \Delta{y}) = \begin{pmatrix}
        \cos(\theta_k \Delta{x}) & -\sin(\theta_k \Delta{x}) & 0 & 0 \\
        \sin(\theta_k \Delta{x}) & \cos(\theta_k \Delta{x}) & 0 & 0 \\
        0 & 0 & \cos(\theta_k \Delta{y}) & -\sin(\theta_k \Delta{y}) \\
        0 & 0 & \sin(\theta_k \Delta{y}) & \cos(\theta_k \Delta{y})
    \end{pmatrix}\enspace,
    $
    }
\end{equation}
where $\theta_k = \frac{1}{10000^{4k/d}},\enspace k\in [1,2,...,d/4]$ encode the index of feature channels.

Compared to the absolute position encoding used in previous methods~\cite{sun2021loftr,wang2022matchformer,chen2022aspanformer,tang2022quadtree,giang2022topicfm,Ni2023PATSPA},
we utilize 2D RoPE to allow the model to focus more on the interaction between features rather than their specific locations, which benefits capturing the context of local features.
Moreover, relative position encoding is more robust to transformations like rotation, translation, and scaling, which is important for matching local features in different views.

\section{More Experiments Results}
\subsection{More Ablation Studies}
In this part, we conduct more ablation studies on the MegaDepth and ScanNet dataset to validate the design choices of our proposed modules.

\paragraph{Position Encoding}
We compare the performance of our 2D RoPE with the sinusoidal position encoding~\cite{Vaswani2017AttentionIA} in Tab.~\ref{tab: sup_abl_rope}.
The results show that using 2D RoPE can achieve better performance than sinusoidal position encoding.
\begin{figure*}[btp]
	\centering
	\includegraphics[width=0.95\textwidth]{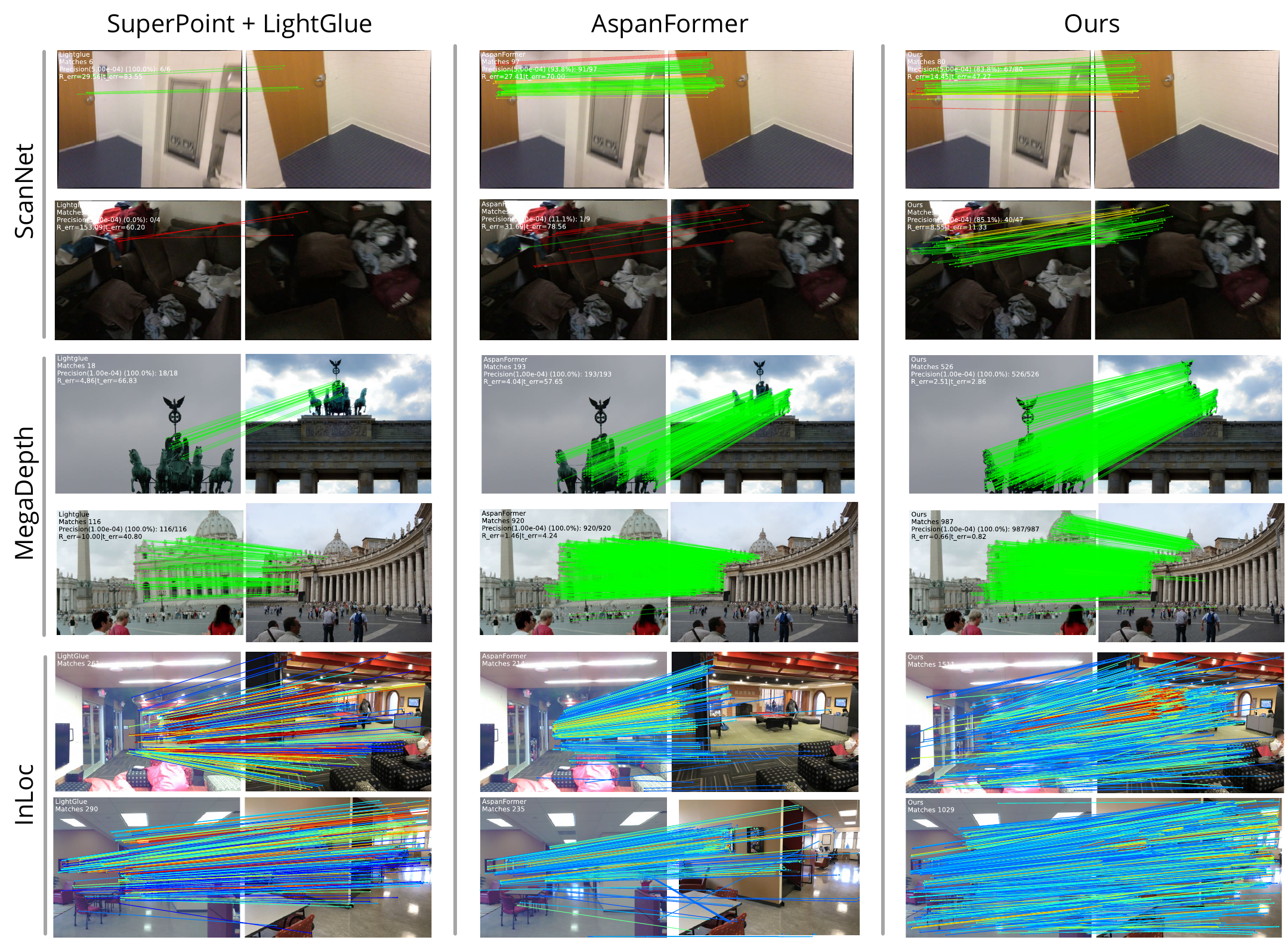}
	\vspace{-0.2 cm}
	\caption{{\textbf{Qualitative Results.}}
	Our method is compared with the sparse matching pipeline SuperPoint~\cite{DeTone2017SuperPointSI}+LightGlue~\cite{lindenberger2023lightglue}, semi-dense matcher AspanFormer~\cite{chen2022aspanformer}. 
	The red color indicates epipolar error beyond $5 \times 10^{-4}$ on ScanNet and $1 \times 10^{-4}$ on MegaDepth (in the normalized image coordinates).
	Since no ground-truth pose is available on InLoc dataset, we color the match with predicted confidence. Red indicates higher confidence and blue for the opposite.
	}
	\label{fig:supqualitative}
	  \vspace{-0.4 cm}
\end{figure*}
\begin{table}[t]
    \centering
    \resizebox{0.75\columnwidth}{!}{
    \setlength\tabcolsep{2pt} %
    \begin{tabular}{lcccc} 
    \toprule
    \multirow{2}{*}{Position Encoding}         & \multicolumn{3}{c}{Pose Estimation AUC} & \multirow{2}{*}{Time~(ms)}\\ 
    \cmidrule(lr){2-4}
        & @5\degree       & @10\degree       & @20\degree \\ 
    \midrule
    RoPE & \textbf{56.4} & \textbf{72.2} & \textbf{83.5}  & 139.2\\
    sinusoidal & 55.5 & 71.5 & 83.1 & \textbf{137.5} \\ 
    \bottomrule
    \end{tabular}
    }
    \vspace{-0.15cm}
    \caption{
    Impact of position encoding on the MegaDepth dataset, where averaged running times for an image pair with high-resolution $1200\times1200$ are reported.
    }
    \label{tab: sup_abl_rope}
    \vspace{-0.35 cm}
    \end{table}

\paragraph{Aggregation Range}
We show the performance of our method with different aggregation range $s$ in Tab.~\ref{tab: sup_abl_agg}.
In our aggregated attention module, we use a $4\times4$ aggregation range to reduce token size before performing attention.
Using a smaller aggregation range leads to more tokens, with slight performance changes but significantly slower matching speed.
This validates the effectiveness of our parameter choice in the aggregation attention module.
\begin{table}[t]
    \centering
    \resizebox{0.75\columnwidth}{!}{
    \setlength\tabcolsep{2pt} %
    \begin{tabular}{ccccc} 
    \toprule
    \multirow{2}{*}{aggregation range}         & \multicolumn{3}{c}{Pose Estimation AUC} & \multirow{2}{*}{Time~(ms)}\\ 
    \cmidrule(lr){2-4}
        & @5\degree       & @10\degree       & @20\degree \\ 
    \midrule
    $s=4$ & \textbf{56.4} & \textbf{72.2} & 83.5  & \textbf{139.2}\\
    $s=2$ & 56.2 & \textbf{72.2} & \textbf{83.6} & 271.1 \\ 
    \bottomrule
    \end{tabular}
    }
    \vspace{-0.15cm}
    \caption{ 
    Impact of aggregation range on the MegaDepth dataset, where averaged running times for an image pair with high-resolution $1200\times1200$ are reported.
    }
    \label{tab: sup_abl_agg}
    \vspace{-0.35 cm}
    \end{table}

\paragraph{Image Resolution}
We test the performance of our method with different image resolutions to show the performance and efficiency changes. 
Results are shown in Tab.~\ref{tab: sup_abl_resolution}.
Compared with the default resolution $1184\times1184$ used in the MegaDepth evaluation, using a larger image size leads to noticeable accuracy improvement with a slower matching speed.
Our method can still achieve competitive performance using low-resolution $640\times640$ images with the fastest speed.
Therefore, our method is pretty robust in image resolution choices for flexible real-world applications.
\begin{table}[t]
    \centering
    \resizebox{0.75\columnwidth}{!}{
    \setlength\tabcolsep{2pt} %
    \begin{tabular}{lcccc} 
    \toprule
    \multirow{2}{*}{Resolution}         & \multicolumn{3}{c}{Pose Estimation AUC} & \multirow{2}{*}{Time~(ms)}\\ 
    \cmidrule(lr){2-4}
        & @5\degree       & @10\degree       & @20\degree \\ 
    \midrule
    $640\times640$ & 51.0 & 67.4 & 79.8  & \textbf{41.7} \\
    $800\times800$ & 53.4 & 70.0 & 81.9 & 58.2 \\ 
    $960\times960$ & 54.7 & 70.7 & 82.4 & 81.8 \\ 
    $1184\times1184$ & \textbf{56.4} & 72.2 & \textbf{83.5} & 139.2 \\ 
    $1408\times1408$ & 56.2 & \textbf{73.1} & 83.4 & 223.9 \\ 
    \bottomrule
    \end{tabular}
    }
    \vspace{-0.15cm}
    \caption{ 
    Impact of test image resolution on the MegaDepth dataset.    
    }
    \label{tab: sup_abl_resolution}
    \end{table}

\paragraph{Linear Attention After Aggregation}
Using linear attention in our aggregated attention module introduces minor efficiency gain on high-resolution but with accuracy dropping as shown in Tab.~\ref{tab: abl_linear_after_agg}.
\begin{table}[t]
    \centering
    % \vspace{-0.7cm}
    \resizebox{1.0\columnwidth}{!}{
    \setlength\tabcolsep{2pt} % default value: 6pt
    % \small % 可以使用更小的字体大小
    \begin{tabular}{ccccccccc} 
    \toprule
    \multirow{2}{*}{Method}         & \multicolumn{3}{c}{MegaDepth Dataset} & \multirow{2}{*}{Time~(ms)} & \multicolumn{3}{c}{ScanNet Dataset} & \multirow{2}{*}{Time~(ms)} \\ 
    \cmidrule(lr){2-4}
    \cmidrule(lr){6-8}
    &       AUC@5$\degree$       & AUC@10$\degree$       & AUC@20$\degree$ & & AUC@5$\degree$       & AUC@10$\degree$       & AUC@20$\degree$  \\ 

    \midrule
    Full  & \textbf{56.4}&\textbf{72.2}&\textbf{83.5}& 139.1& \textbf{19.2}&\textbf{37.0}&\textbf{53.6} & \textbf{34.4} \\
    Linear& 54.1& 70.3 & 82.1 & \textbf{132.7}  & 16.8 & 33.2 & 49.0 & 36.9 \\

    \bottomrule
    \end{tabular}
    }

    \vspace{-0.15cm}
    \caption{
    Impact of linear attention after aggregation on the MegaDepth and ScanNet dataset, where the resolution are $1200\times1200$ and $640\times480$, respectively.
    }
    \label{tab: abl_linear_after_agg}
    \vspace{-0.35 cm}
    \end{table}

\paragraph{Additional ablation studies on the ScanNet dataset}
We further repeat the ablation studies in the main paper and conduct additional ablation studies on the ScanNet dataset to validate the design choices of our proposed modules.
Results are shown in Tab.~\ref{tab: abl_scannet}.
\begin{table}[t]
    \centering
    % \vspace{-0.3cm}
    \resizebox{1.0\columnwidth}{!}{
    \setlength\tabcolsep{2pt} % default value: 6pt
    \begin{tabular}{lcccc} 
    \toprule
    %  & @5\degree & @10\degree & @20\degree & Time~(ms)\\ 
    \multirow{2}{*}{Method}         & \multicolumn{3}{c}{Pose Estimation AUC} & \multirow{2}{*}{Time~(ms)}\\ 
    \cmidrule(lr){2-4}
        & @5\degree       & @10\degree       & @20\degree \\ 

    \midrule
    Ours Full & \textbf{19.2} & \textbf{37.0} & \textbf{53.6}  & 34.4\\
    1) Ours Optimal~(w/o dual-softmax) & 17.4 & 34.4 & 51.2 & \textbf{27.0} \\ 
    2) Replace Agg. Attention to LoFTR's Trans. & 17.1&33.2&49.4 & 41.3\\
    3) Replace two-stage refine. to LoFTR's refine. & 18.1 & 35.8 & 52.4 & 31.8 \\
    4) No second-stage refinement & 18.8 & 36.7 & 53.4 & 32.2\\
    5) Replace RepVGG with ResNet & 18.6&36.3&52.8 & 38.1\\
    6) Both Conv in Agg. Attention &18.6&35.8&52.5&34.4 \\
    7) Both Pool in Agg. Attention &18.3 & 35.2& 51.7&34.1 \\
    8) Upsample after FFN &17.3 & 34.6& 51.4&32.6 \\
    \bottomrule
    \end{tabular}
    }
    \vspace{-0.15cm}
    \caption{
    The components of our method are ablated on the ScanNet dataset again for a comprehensive understanding of our method, where averaged running times for an image pair with resolution $640\times480$ are reported.
    }
    \label{tab: abl_scannet}
    \vspace{-0.35 cm}
    \end{table}

\subsection{More Qualitative Results}
More qualitative results on the ScanNet dataset, InLoc dataset, and MegaDepth dataset are shown in Fig.~\ref{fig:supqualitative}.

\subsection{Additional Results on other RANSAC setting}
LightGlue uses a RANSAC setting different from other baseline papers~\cite{sarlin20superglue,sun2021loftr,tang2022quadtree,wang2022matchformer,giang2022topicfm,chen2022aspanformer,edstedt2023dkm,edstedt2023roma} in relative pose estimation evaluations on MegaDepth. 
We further conduct experiments following LightGlue's setting (OpenCV RANSAC~\cite{Fischler1981RandomSC} and LO-RANSAC~\cite{PoseLib} with carefully tuned RANSAC inlier thresholds), as shown in Tab.~\ref{tab:exp tuned_RANSAC}.
Using the naive RANSAC method, the performance gap between ours and LightGlue becomes larger after RANSAC threshold tuning~(compared with our untuned results in Tab.~\ref{tab:exp relativepose}).
This demonstrates that our method can achieve significantly better accuracy without depending on the sophisticated modern RANSAC method, thereby revealing its superior match quality.
Using the stronger outlier filter LO-RANSAC, the accuracy of all methods is boosted and our method consistently achieves better performance than LightGlue, especially on the AUC$5\degree$ metric.
\begin{table}[t]
    % \vspace{-0.4 cm}
    \centering
    \resizebox{0.45\textwidth}{!}{
    \setlength\tabcolsep{4pt} %
    \begin{tabular}{ccccccc} 
    \toprule
    \multirow{2}{*}{Method}         & \multicolumn{3}{c}{RANSAC}             & \multicolumn{3}{c}{LO-RANSAC} \\ 
    \cmidrule(lr){2-4}
    \cmidrule(lr){5-7}
                  & AUC@5$\degree$       & AUC@10$\degree$       & AUC@20$\degree$  & AUC@5$\degree$       & AUC@10$\degree$       & AUC@20$\degree$  \\ 
    \midrule
    LightGlue & 49.9&67.0&80.1 & 66.8&79.3&87.9  \\
    AspanFormer & 58.3&73.3&\textbf{84.2} & 69.4&\textbf{81.1}&\textbf{88.9}  \\
    Ours & \textbf{58.4}&\textbf{73.4}&\textbf{84.2}&\textbf{69.5}&80.9&88.8 \\
    \bottomrule
    \end{tabular}
    }
    \vspace{0.15cm}
    \caption{Results of Relative Pose Estimation on MegaDepth Dataset following LightGlue's setting.
    The AUC of pose error at different thresholds is presented. 
    }
    \label{tab:exp tuned_RANSAC}
    \vspace{-0.35 cm}
    \end{table}

\section{Details About Timing}
The running times evaluated in the paper are averaged over all pairs in the test dataset with a warm-up of 50 pairs for accurate measurement.
All the methods are tested on a single NVIDIA RTX 3090 GPU with 14 cores of Intel Xeon Gold 6330 CPU.

We further report each part running time of our method in Tab.~\ref{tab: sup_timing}, where both full and optimized models are shown.
We noticed that skipping the dual-softmax of Coarse Matching in the optimized model can significantly reduce the running time.
What's more, with the benefit of Mixed-Precision, the running time of feature extraction can be further reduced.
\begin{table}[t]
    \centering
    \resizebox{0.75\columnwidth}{!}{
    \setlength\tabcolsep{4pt} %
    \begin{tabular}{lcc} 
    \toprule
    \multirow{2}{*}{Process}         & \multicolumn{2}{c}{Time(ms)}  \\ %
    \cmidrule(lr){2-3}
        & Full       & Optimized \\ 
    \midrule
    Total & 40.1 & 27.0 \\
    \midrule
    Feature Backbone & 9.1 & 5.8 \\
    Coarse Feature Transformation & 11.7 & 12.9 \\
    Coarse Matching & 8.3 & 1.7 \\
    Fine Feature Fusion  & 8.0 & 4.8 \\
    Two-Stage Refinement & 3.0 & 2.0 \\
    \bottomrule
    \end{tabular}
    }
    \vspace{-0.15cm}
    \caption{ 
    Time cost for an image pair of $640\times480$ on the ScanNet dataset. 
    The optimized model uses Mixed-Precision numerical accuracy and drops the dual-softmax operator in the coarse matching phase.
    }
    \label{tab: sup_timing}
    \vspace{-0.35 cm}
    \end{table}

\begin{table}[t]
    \centering
    \resizebox{0.75\columnwidth}{!}{
    \setlength\tabcolsep{4pt} %
    \begin{tabular}{lccc} 
    \toprule
    \multirow{2}{*}{Method}         & \multicolumn{2}{c}{Time(ms)} & \multirow{2}{*}{\#Matches}  \\ %
    \cmidrule(lr){2-3}
        & Matching       & RANSAC &  \\ 
    \midrule
    SP + SG & 43.6 & 0.53 & 487 \\
    DRC-Net & 143.9 & 2.78 & 1019 \\ 
    LoFTR & 76.2 & 1.59 & 995 \\ 
    Ours & 45.9/38.6 & 1.39 & 997 \\
    \bottomrule
    \end{tabular}
    }
    \vspace{-0.15cm}
    \caption{
    Running times of different methods on HPatches dataset. 
    All images are resized so that their short edge equals 480 pixels following SuperGlue~\cite{sarlin20superglue} and LoFTR~\cite{sun2021loftr}.
    For Ours, the running times of the model using FP32/Mixed-Precision numerical precisions are shown.
    }
    \label{tab: sup_hpatches_timing}
    \vspace{-0.35 cm}
    \end{table}
\begin{table}[t]
    \centering
    \resizebox{0.6\columnwidth}{!}{
    \setlength\tabcolsep{4pt} %
    \begin{tabular}{lcc} 
    \toprule
    \multirow{2}{*}{Method}         & \multicolumn{2}{c}{Time(ms)}  \\ %
    \cmidrule(lr){2-3}
        & Aachen       & InLoc \\ 
    \midrule
    SP+SG       & 55.9& 83.3\\
    LoFTR       & 83.2& 147.6\\
    TopicFM     & 66.0& 119.6\\
    PATS        & 315.8& 1148.0\\
    AspanFormer & 95.4& 164.5\\
    Ours        & \textbf{40.6/25.5}& \textbf{82.0/46.3} \\
    \bottomrule
    \end{tabular}
    }
    \vspace{-0.15cm}
    \caption{
    Running times of different methods on Aachen and InLoc dataset.
    To measure the running time, we sample 818 and 356 pairs of images from the NetVLAD~\cite{Arandjelovi2015NetVLADCA}'s retrieval results for Aachen and InLoc, respectively.
    All images are resized so that their long edge equals 1024 pixels following HLoc~\cite{Sarlin2018FromCT}.
    For Ours, the running times of the model using FP32/Mixed-Precision numerical precisions are shown.
    }
    \label{tab: sup_loc_timing}
    \vspace{-0.35 cm}
    \end{table}
The latency on $3$ datasets included in \cref{tab:exp hpatches,tab:exp inloc,tab:exp aachen} are shown in Tab.~\ref{tab: sup_hpatches_timing} 
and Tab.~\ref{tab: sup_loc_timing}, where conclusion and speed rankings are the same as indicated in Tab.~\ref{tab:exp relativepose}.
We further show RANSAC time on HPatches dataset. Both matching and RANSAC latency of our method are smaller than LoFTR with a similar number of matches.

\section{Limitations and Future Works}
We find that our method may fail when strong repetitive structures exist, such as matching an image pair that depicts different scenes containing the same chair.
We think this may be due to the current model focusing more on local features for accurate matching, where global semantic context is lacking.
Therefore, the mechanism of high-level contexts can be added to the model for performance improvement on ambiguous scenes.
Moreover, we believe the efficiency of our method can be further improved by adopting the early stop strategy of LightGlue~\cite{lindenberger2023lightglue} since the contribution of the proposed efficient aggregation module is orthogonal to it.

\end{document}